%% file: main.tex
\documentclass[runningheads]{llncs}
\usepackage{graphicx}

\usepackage[dvipsnames]{xcolor}
\usepackage{kvoptions}
\usepackage{xcolor-material}

\usepackage[caption=false]{subfig}

\usepackage{tikz}
\usepackage{comment} 
\usepackage{amsmath,amssymb} 
\usepackage{color}

\usepackage{appendix}
\makeatletter
\newcommand{\@chapapp}{\relax}%
\makeatother

\usepackage{multirow}
\usepackage{booktabs}

\usepackage{pgfplots}

\newcommand{\E}{\mathcal{E}}
\newcommand{\Surf}{\mathcal{S}}
\newcommand{\M}{\mathcal{M}}
\newcommand{\xy}{(x,y)}

\newcommand{\R}{\mathbb{R}}

\usepackage{array}
\newcommand{\PreserveBackslash}[1]{\let\temp=\\#1\let\\=\temp}
\newcolumntype{C}[1]{>{\PreserveBackslash\centering}p{#1}}
\newcolumntype{R}[1]{>{\PreserveBackslash\raggedleft}p{#1}}
\newcolumntype{L}[1]{>{\PreserveBackslash\raggedright}p{#1}}

\begin{document}
\pagestyle{headings}
\mainmatter
\def\ECCVSubNumber{3398}  

\title{A Differentiable Recurrent Surface for Asynchronous Event-Based Data} 

\titlerunning{A Differentiable Recurrent Surface for Asynchronous Event-Based Data}
%
\author{
Marco Cannici \and
Marco Ciccone \and
Andrea Romanoni~\thanks{Work done prior to Amazon involvement of the author and does not reflect views of the Amazon company.} \and
Matteo Matteucci
}
\authorrunning{M. Cannici et al.}
%
\institute{Politecnico di Milano, Italy \\
\email{\{marco.cannici,marco.ciccone,andrea.romanoni,matteo.matteucci\}@polimi.it}}
\maketitle

\begin{abstract}
Dynamic Vision Sensors (DVSs) asynchronously stream events in correspondence of pixels subject to brightness changes. Differently from classic vision devices, they produce a sparse representation of the scene. Therefore, to apply standard computer vision algorithms, events need to be integrated into a frame or event-surface. This is usually attained through hand-crafted grids that reconstruct the frame using ad-hoc heuristics. In this paper, we propose Matrix-LSTM, a grid of Long Short-Term Memory (LSTM) cells that efficiently process events and learn end-to-end task-dependent event-surfaces. Compared to existing reconstruction approaches, our learned event-surface shows good flexibility and expressiveness on optical flow estimation on the MVSEC benchmark and it improves the state-of-the-art of event-based object classification on the N-Cars dataset.
\end{abstract}

\section{Introduction}
\label{sec:intro}

Event-based cameras, such as dynamic vision sensors (DVSs)~\cite{lichtsteiner2008128,serrano2013128,berner2013240,posch2014retinomorphic}, are bio-inspired devices that attempt to emulate the efficient data-driven communication mechanisms of the brain. Unlike conventional frame-based active pixel sensors (APS), which capture the scene at a predefined and constant frame-rate, these devices are composed of independent pixels that output sequences of asynchronous events, efficiently encoding pixel-level brightness changes caused by moving objects. This results in a sensor having a very high dynamic range ($>120$ dB) and high temporal resolution (in the order of microseconds), matched with low power consumption and minimal delay. All these characteristics are key features in challenging scenarios involving fast movements (e.g., drones or moving cars), and abrupt brightness changes (e.g., when exiting a dark tunnel in a car). 
However, novel methods and hardware architectures need to be specifically designed to exploit these advantages and leverage their potential in complex tasks. Event-cameras only provide a timed sequence of changes that is not directly compatible with computer vision systems which typically work on frames. 

Driven by the great success of frame-based deep learning architectures, that learn representations directly from standard APS signals, research in event-based processing is now focusing on how to effectively aggregate event information in grid-based representations which can be directly used, for instance, by convolutional deep learning models. Nevertheless, finding the best mechanism to extract information from event streams is not trivial.
Multiple solutions have indeed emerged during the past few years, mostly employing hand-crafted mechanisms to accumulate events. Examples of such representations are mechanisms relying on exponential~\cite{cohen2016event,lagorce2016hots,sironi2018hats} and linear~\cite{cohen2016event,cannici2019asynchronous} decays, ``event-surfaces'' storing the timestamp of the last received event in each pixel and extensions of such mechanism making use of memory cells~\cite{sironi2018hats} and voxel-grids~\cite{rebecq2019events,Zhu2018ev-flownet}. 

Only very recently deep learning techniques have been applied to learn such surfaces in a data-driven manner~\cite{gehrig2019end}. In this paper, we focus on this recent trend in event-based processing, and propose a mechanism to efficiently apply a Long Short-Term Memory (LSTM) network~\cite{hochreiter1997long} as a convolutional filter over the 2D stream of events in order to accumulate pixel information through time and build 2D event representations. The reconstruction mechanism is end-to-end differentiable, meaning that it can be jointly trained with state-of-the-art frame-based architectures to learn event-surfaces specifically tailored for the task at hand. Most importantly, the mechanism specifically focuses on preserving sparsity during computation, enabling the reconstruction process to only focus on pixels receiving events and  without requiring events to be densified in a dense tensor during the intermediate feature extraction steps, process that is otherwise necessary when applying standard computer vision approaches, such as ConvLSTM \cite{shi2015convolutional}, in most of the cases.

Substituting hand-crafted event-surfaces with our trainable layer in state-of-the-art architectures improves their performance substantially without requiring particular effort in hyper-parameter tuning, enabling researchers to exploit event information effectively.
The contributions of the paper are summarized as follows:

\begin{itemize}
    \item We propose Matrix-LSTM, a task-independent mechanism to extract grid-like event representations from asynchronous streams of events. The framework is end-to-end differentiable, it can be used as input of any existing frame-based state-of-the-art architecture and jointly trained to extract the best representation from the events. 
    
    \item Replacing input representations with a Matrix-LSTM layer in existing architectures, we show that it improves the state-of-the-art on event-based object classification on N-CARS~\cite{sironi2018hats} by $3.3\%$ and performs better than hand-crafted features on N-Caltech101~\cite{orchard2015converting}. Finally, it improves optical flow estimation on the MVSEC benchmark~\cite{zhu2018multivehicle} up to $30.76\%$ over hand-crafted features~\cite{zhu2018multivehicle} and up to $23.07\%$ over end-to-end differentiable ones~\cite{gehrig2019end}.
    
    \item We developed custom CUDA kernels, both in PyTorch~\cite{steiner2019pytorch} and TensorFlow~\cite{abadi2019tensorflow}, to efficiently aggregate events by position and perform a convolution-like operation on the stream of events using an LSTM as a convolutional filter~\footnote{Code available at \url{https://marcocannici.github.io/matrixlstm}}.
\end{itemize}

\section{Related Work}
\label{sec:related-work}
Event cameras provide outstanding advantages over ordinary devices in terms of time resolution and dynamic range. However, their potentialities are still unlocked, mainly due to the difficulty of building good representations from a sparser, asynchronous and much more rough source of information compared to frame-based data. In this section, we give a brief overview of related works, focusing on representations for event-based data and highlighting the differences and similarities with our work. We refer the reader to~\cite{gallego2019event} for a thorough overview.


\paragraph{\textbf{Hand-crafted representations.}}
Several hand-crafted event representations have been proposed over the years, ranging from biologically inspired, such as those used in Spiking Neural Networks~\cite{maass1997networks}, to more structured ones.
Recently, the concept of \emph{time-surface} was introduced~\cite{lagorce2016hots,maqueda2018event}, in which 2D surfaces are obtained by keeping track of the timestamp of the last event occurred in each location and by associating each event with features computed applying exponential kernels on the surface. An extension of these methods, called HATS~\cite{sironi2018hats}, employs memory cells that retain temporal information from past events. Instead of building the surface using just the last event, too sensitive to noise, HATS uses a fixed-length memory. Histograms are then extracted from the surface and a SVM classifier is finally used for prediction. The use of a memory to compute the event-surface closely relates HATS with the solution presented in this paper. Crucially, the accumulation procedure employed in HATS is hand-crafted, while our work is end-to-end trainable thanks to a grid of LSTM cells~\cite{hochreiter1997long}, which enable to learn a better accumulation strategy directly from data.

In~\cite{Zhu2018ev-flownet}, the authors propose the EV-FlowNet network for optical flow estimation together with a new time-surface variant. Events of different polarities are kept separate to build a four-channel grid containing the number of events occurred in each location besides temporal information. A similar representation has also been used in~\cite{ye2018unsupervised}. To improve the temporal resolution of such representations, \cite{zhu2019unsupervised} suggests to discretize time into consecutive bins and accumulate events into a voxel-grid through a linearly weighted accumulation similar to bilinear interpolation. A similar time discretization has also been used in Events-to-Video~\cite{rebecq2019events}, where the event representation is used within a recurrent-convolutional architecture to produce realistic video reconstructions of event sequences. Despite being slower, the quality of reconstructed frames closely resembles actual gray-scale frames, allowing the method to take full advantage from transferring feature representations trained on natural images.

\paragraph{\textbf{End-to-end representations.}}
Most closely related to the current work, \cite{gehrig2019end} learns a dense representation end-to-end directly from raw events. A multi-layer perceptron (MLP) is used to implement a trilinear filter that produces a voxel-grid of temporal features. The event time information of each event is encoded using the MLP network and the value obtained from events occurring in the same spatial location are summed up together to build the final feature. A look-up table is then used, after training, to speed-up the procedure. Events are processed independently as elements of a set, disregarding their sequentiality and preventing the network to modulate the information based on previous events. Our method, instead, by leveraging the memory mechanism of LSTM cells, can integrate information conditioned on the current state and can decide how much each event is relevant to perform the task, and how much information to retain from past events.
A recent trend in event-based processing is studying mechanisms that do not require to construct intermediate explicit dense representations to perform the task at hand~\cite{bi2019graph,sekikawa2019eventnet,wang2019space}. Among these, \cite{neil2016phased} uses a variant of the LSTM network, called PhasedLSTM, to learn the precise timings of events. While it integrates the events sequentially as in our work, PhasedLSTM employs a single cell on the entire stream of events and can be used only on very simple tasks~\cite{cannici2019attention}. The model, indeed, does not maintain the input spatial structure and condenses the 2D stream of events into a single feature vector, preventing the network to be used as input to standard CNNs. 
Finally, although it has never been adopted with event-based cameras, we also mention here the ConvLSTM \cite{shi2015convolutional} network, a convolutional variant of the LSTM that has previously been applied on several end-to-end prediction tasks. Despite its similarity with our method, since both implement the notion of convolution to LSTM cells, ConvLSTM is not straightforward to apply to sparse event-based streams and requires the input to be densified into frames before processing. This involves building very sparse frames of simultaneous events, mostly filled with padding, or dense frames containing uncorrelated events. Our formulation, instead, preserves sparsity during computation and does not require events to be densified, even when large receptive fields are considered.

\section{Method}
\label{sec:method}

Event-based cameras are vision sensors composed of pixels able to work independently. Each pixel has its own exposure time and it is free to fire independently by producing an event as soon as it detects a significant change in brightness. Unlike conventional devices, no rolling shutter is used, instead, an asynchronous stream of events is generated describing what has changed in the scene. Each event $e_i$ is a tuple $e_i = (x_i, y_i, t_i, p_i)$ specifying the time $t_i$, the location $\xy_i$ (within a $H \times W$ space) and the polarity $p_i \in \{-1, 1\}$ of the change (brightness increase or decrease). Therefore, given a time interval $\tau$ (i.e., the sample length), the set of events produced by the camera can be described as a sequence $\E = \{ (x_i, y_i, t_i, p_i) \mid t_i \in \tau \}$, ordered by the event timestamp. In principle, multiple events could be generated at the same timestamp. However, the grid representation of the events at a fixed timestamp $t$ is likely to be very sparse, hence, an integrating procedure is necessary to reconstruct a dense representation $\Surf_{\E}$ before being processed by conventional frame-based algorithms.

Note that, in this work, we do not aim to reconstruct a frame that resembles the actual scene, such as a grey-scale or RGB image~\cite{rebecq2019events,scheerlinck2019ced}, but instead to extract task-aware features regardless of their appearance. In the following, \textit{``surface''},  \textit{``reconstruction''} and  \textit{``representation''} are used with this meaning. 

\begin{figure}[t]
    \centering
    \includegraphics[width=\linewidth]{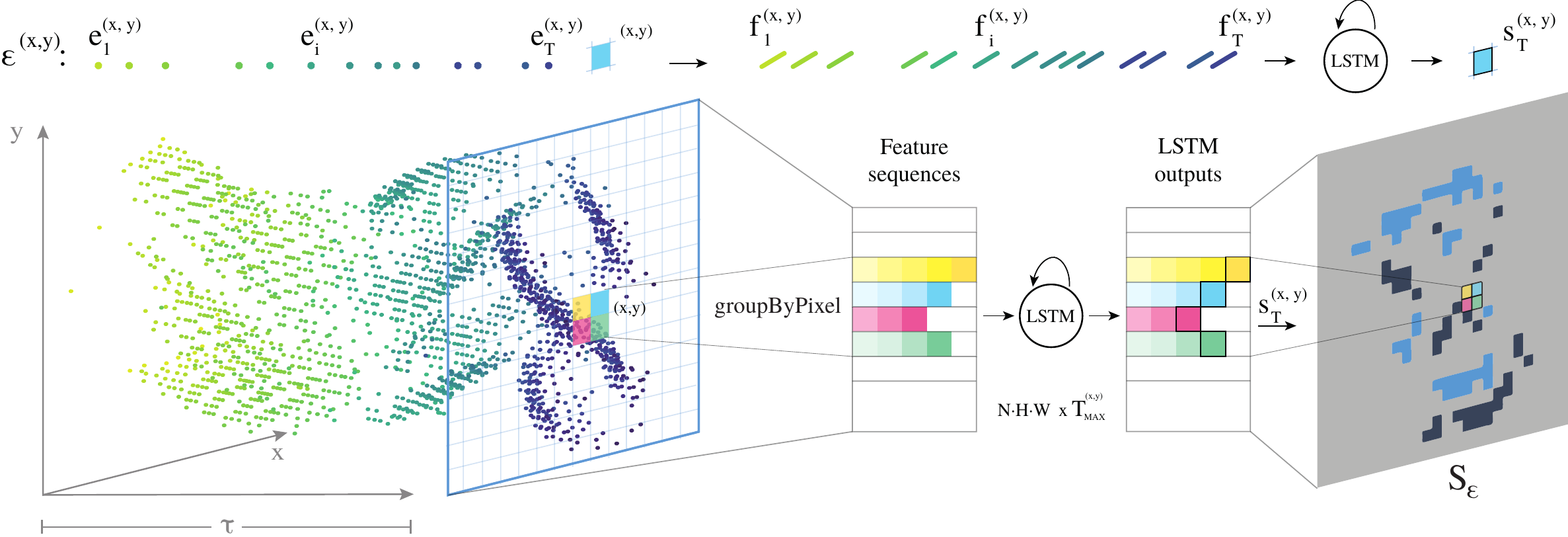}
    
    \caption{Overview of Matrix-LSTM (figure adapted from \cite{neil2016phased}). Events in each pixel are first associated to a set of features $f^{\xy}_i$, and then processed by the LSTM. The last output,  $s^{\xy}_{T}$, is finally used to construct $\Surf_{\E}$. \emph{GroupByPixel} is shown here on a single sample ($N = 1$) highlighting a $2\times2$ pixel region. Colors refer to pixel locations while intensity indicates time. For clarity, the features dimension is not shown in the figure}
    
    \label{multifig:matrixlstm_groupbypixel}
\end{figure}

\subsection{Matrix-LSTM}
\label{sec:matrixlstm}
Analogously to~\cite{gehrig2019end}, our goal is to learn end-to-end a fully parametric mapping $\M: \E \to \Surf_{\E} \in \R^{H \times W \times C}$, between the event sequence and the corresponding dense representation, providing the best features for the task to optimize.

In this work, we propose to implement $\M$ as an $H \times W$ \textit{matrix of LSTM cells}~\cite{hochreiter1997long} (see Figure \ref{multifig:matrixlstm_groupbypixel}). Let's define the ordered sequence of events $\E^{\xy}$ produced by the pixel $\xy$ during interval $\tau$ as $\E^{\xy} = \{ (x_i, y_i, t_i, p_i) \mid t_i \in \tau, x_i=x, y_i=y\} \subset \E$, and its length as $T^{\xy} = \vert \E^{\xy} \vert$, which may potentially be different for each location $\xy$. A set of features $f^{\xy}_i \in \R^F$ is first computed for each event occurring at location $\xy$, typically the polarity and one or multiple temporal features (see Section~\ref{sec:evaluation}). At each location $\xy$, an $LSTM^{\xy}$ cell then processes these features asynchronously, keeping track of the current integration state and condensing all events into a single output vector $s^{\xy} \in \R^C$. 
In particular, at each time $t$, the $LSTM^{\xy}$ cell produces an intermediate representation $s^{\xy}_{t}$. Once all the events are processed, the last output of the LSTM cell compresses the dynamics of the entire sequence $\E^{\xy}$ into a fixed-length vector $s^{\xy}_{T}$ that can be used as pixel feature (here we dropped the superscript $^{\xy}$ from $T$ for readability). The final surface $\Surf_{\E}$ is finally built by collecting all LSTMs final outputs $s^{\xy}_{T}$ into a dense tensor of shape $H \times W \times C$. 
A fixed all-zeros output is used where the set of events $\E^{\xy}$ is empty.

\paragraph{\textbf{Temporal bins.}} Taking inspiration from previous methods~\cite{gehrig2019end,rebecq2019events,zhu2019unsupervised} that discretize time into temporal bins, we also propose a variant of Matrix-LSTM that operates on successive time windows.
Given a fixed number of bins $B$, the original event sequence is split into $B$ consecutive windows $\E_{\tau_1}, \E_{\tau_2}, ..., \E_{\tau_B}$. Each sequence is processed independently, i.e., the output of each LSTM at the end of each interval is used to construct a surface $\Surf_{\E_b}$ and the LSTMs state is re-initialized before the next sub-sequence starts. This gives rise of $B$ different reconstructions $\Surf_{\E_b}$ that are concatenated to form the final surface $\Surf_{\E} \in \R^{H \times W \times B \cdot C}$. In this formulation, the LSTM input features $f^{\xy}_i$ usually contain both global temporal features (i.e., w.r.t. the original uncut sequence) and relative features (i.e., the event position in the sub-sequence). 
Although LSTMs should be able to retain memory over very long periods, we found that discretizing time into intervals helps 
, especially in tasks requiring precise time information such as optical flow estimation (see Section~\ref{par:opticalflow-results}). A self-attention module~\cite{hu2018squeeze} is then optionally applied on the reconstructed surface to correlate intervals (see Section~\ref{par:classification-results}). 

\paragraph{\textbf{Parameters sharing.}}
Inspired by the convolution operation defined on images, we designed Matrix-LSTM to enjoy translation invariance.
This is implemented by sharing the parameters across all the LSTM cells, as in a convolutional kernel.
Sharing parameters not only drastically reduces the number of parameters in the network, but it also allows us to transfer a learned transformation to higher or lower resolutions as in fully-convolutional networks~\cite{Long2014Nov}.


We highlight that such an interpretation of the Matrix-LSTM functioning also fits the framework proposed in~\cite{gehrig2019end}, in which popular event densification mechanisms are rephrased as kernel convolutions on the \textit{event field}, i.e., a discretized four-dimensional manifold spanning $x$ and $y$, and the time and polarity dimensions. We finally report that this formulation is equivalent to a $1 \times 1$ ConvLSTM \cite{shi2015convolutional} applied on a dense tensor where events are stacked in pixel locations by arrival order. However, as reported in Section \ref{sec:convlstm_comparison}, this formulation has better space and time performance on sparse event sequences. Moreover, in the next section, an extension to larger receptive fields with better accuracy performance on asynchronous event data compared to ConvLSTM, is also proposed.

\paragraph{\textbf{Receptive field size.}}
As in a conventional convolution operation, Matrix-LSTM can be convolved on the input space using different strides and kernel dimensions. In particular, given a receptive field of size $K_H \times K_W$, each LSTM cell processes a local neighborhood of asynchronous events $\E^{\xy} = \{ (x_i, y_i, t_i, p_i) \mid t_i \in \tau, |x-x_i| < K_W - 1, |y-y_i| < K_H - 1 \}$. Events features are computed as in the original formulation, however, an additional coordinate feature $(p_x,p_y)$ is also added specifying the relative position of each event within the receptive field. Coordinate features are range-normalized in such a way that an event occurring in the top-left pixel of the receptive field has feature $(0,0)$, whereas one occurring in the bottom-right position has features $(1,1)$. Events belonging to multiple receptive fields (e.g., when the LSTM is convolved with a stride $1 \times 1$ and receptive field greater then $1 \times 1$) are processed multiple times, independently.

\paragraph{\textbf{Implementation.}}
\label{sec:matrixlstm-implementation}
The convolution-like operation described in the previous section can be implemented efficiently by means of two carefully designed event grouping operations. Rather than replicating the LSTM unit multiple times on each spatial location, a single recurrent unit is applied over different $\E^{\xy}$ sequences in parallel. 
This requires a reshape operation, i.e., \textit{groupByPixel}, that splits events based on their pixel location maintaining the events relative ordering within each sub-sequence. A similar procedure, i.e., \textit{groupByTime}, is employed to efficiently split events into consecutive temporal windows without making use of expensive masking operations. An example of the \textit{groupByPixel} operation is provided in Figure \ref{multifig:matrixlstm_groupbypixel} while implementation details of both operations, implemented as custom CUDA kernels, are provided in the supplementary materials. We finally highlight that these operations are not specific to Matrix-LSTM, since grouping events by pixel index is a common operation in event-based processing, and could indeed benefit other implementations making use of GPUs.

\section{Evaluation}
\label{sec:evaluation}
We test the proposed mechanism on two different tasks: object classification (see Section~\ref{sec:classification}) and optical flow estimation (see Section~\ref{sec:opticalflow}), where the network is required to extract effective temporal features. We evaluated the goodness of Matrix-LSTM features indirectly: a state-of-the-art architecture is taken as a reference and the proposed method is evaluated in terms of the gain in performance obtained by replacing the network representation with a Matrix-LSTM.

\subsection{Object classification}
\label{sec:classification}
We evaluated the model on the classification task using two publicly available event-based collections, namely the N-Cars~\cite{sironi2018hats} and the N-Caltech101~\cite{orchard2015converting} datasets, which represent to date the most complex benchmarks for event-based classification. N-Cars is a collection of urban scenes recordings (lasting $100$ms each) captured with a DVS sensor and showing two object categories: cars and urban background. The dataset comes already split into $7,940$ car and $7,482$ background training samples, and $4,396$ car and $4,211$ background testing samples. The N-Caltech101 collection is an event-based conversion of the popular Caltech-101~\cite{feifei2016one} dataset obtained by moving an event-based camera in front of a still monitor showing one of the original RGB images. Like the original version, the dataset contains objects from $101$ classes distributed amongst $8,246$ samples.

\paragraph{\textbf{Network Architectures.}}
\label{par:classification-network}
We used two network configurations to test Matrix-LSTM on both datasets, namely the classifier used in Events-to-Video~\cite{rebecq2019events}, and the one used to evaluate the EST~\cite{gehrig2019end} reconstruction. Both are based on ResNet~\cite{he2016deep} backbones and pre-trained on ImageNet~\cite{deng2009imagenet}. Events-to-Video~\cite{rebecq2019events} uses a ResNet18 configuration maintaining the first 3 channels convolution (since reconstructed images are RGB) while adding an extra fully-connected layer to account for the different number of classes in both N-Calthec101 and N-Cars (we refer to this configuration as \textit{ResNet--Ev2Vid}). EST~\cite{gehrig2019end} instead uses a ResNet34 backbone and replaces both the first and last layers respectively, with a convolution matching the input features, and a fully-connected layer with the proper number of neurons (we refer to this configuration as \textit{ResNet--EST}). 

To perform a fair comparison we replicated the two settings, using the same number of channels in the event representation (although we also tried different channel values) and data augmentation procedures (random horizontal flips and crops of $224 \times 224$ pixels). We perform early stopping on a validation set in all experiments, using $20\%$ of the training on N-Cars and using the splits provided by the EST official code repository~\cite{uzhrpg2019github} for N-Caltech101. ADAM~\cite{adam2019kingma} was used as optimizer for all experiments with a learning rate of $10^{-4}$. Finally, we use a batch size of $64$ and a constant learning rate on N-Cars in both configurations. On N-Caltech101, instead, we use a batch size of $16$ while decaying the learning rate by a factor of $0.8$ after each epoch when testing on \textit{ResNet--Ev2Vid}, and a batch size of $100$ with no decay with the \textit{ResNet--EST} setup. Finally, to perform a robust evaluation, we compute the mean and standard deviation values using five different seeds in all the experiments reported in this section.

\subsubsection{Results}
\label{par:classification-results}

\begin{table}[t]
    \caption{Results on N-Cars: \textbf{(a)} ResNet18--Ev2Vid, variable time encoding, and normalization; \textbf{(b)} ResNet18--EST, variable time encoding and number of bins}
    \label{multitable:NCARS_Ev2Vid_Norm_Grads_EST_Bins}
    
    \subfloat[\label{table:NCARS_Ev2Vid_ResNet18_TimeEncodings_Norm_Grads}]{\begin{minipage}{.46\textwidth}
        \centering
        \resizebox{0.98\linewidth}{!}{%
        \begin{tabular}{c|ccc}
          \begin{tabular}[c]{@{}c@{}}ResNet \\Norm\end{tabular}        & ts absolute      & ts relative      & delay relative  \\ \hline
          \checkmark  & $95.22\pm0.41\%$ & $94.77\pm1.01\%$ & $95.40\pm0.59\%$ \\
                      & $\mathbf{95.75\pm0.27\%}$ & $95.32\pm0.85\%$ & $\mathbf{95.80\pm0.53\%}$ \\
        \end{tabular}%
        }
    \end{minipage}}
    \subfloat[\label{table:NCARS_EST_ResNet18_TimeEncodings_Bins}]{\begin{minipage}{.54\textwidth}
        \centering
        \resizebox{0.98\linewidth}{!}{%
        \begin{tabular}{cc|ccc}
         &                                          & $1$ bin          & $2$ bins          & $9$ bins \\ \hline 
        \multirow{2}{*}{delay} & glob+loc    & -                & $92.68\pm1.23\%$  & $92.32\pm1.02\%$ \\
         & local                                    & $92.64\pm1.21\%$ & $92.35\pm0.83\%$  & $92.67\pm0.90\%$ \\ \hline
        \multirow{2}{*}{ts}    & ts glob+loc    & -                & $\mathbf{93.46\pm0.84\%}$  & $\mathbf{93.21\pm0.49\%}$ \\
         & local                                    & $92.65\pm0.78\%$ & $92.75\pm1.38\%$  & $93.12\pm0.68\%$
        \end{tabular}%
        }
    \end{minipage}}
\end{table}

The empirical evaluation is organized as it follows for both \textit{ResNet--Ev2Vid} and \textit{ResNet--EST}. We always perform hyper-parameters search using ResNet18 on N-Cars, being faster to train and thus allowing to explore a larger parameter space. We then select the best configuration to train the remaining architectures, i.e., ResNet34 on N-Cars and both variants on N-Caltech101.

\paragraph{\textbf{Matrix-LSTM + ResNet-Ev2Vid.}} We start out with the \textit{ResNet--Ev2Vid} baseline (setting up the Matrix-LSTM to output $3$ channels) by identifying the optimal time feature to provide as input to the LSTM, as reported in Table~\ref{multitable:NCARS_Ev2Vid_Norm_Grads_EST_Bins}\subref*{table:NCARS_Ev2Vid_ResNet18_TimeEncodings_Norm_Grads}.
We distinguish between \textit{ts} and \textit{delay} features and between \textit{absolute} and \textit{relative} scope. The first distinction refers to the type of time encoding, i.e., the timestamp of each event in the case of \textit{ts} feature, or the delay between an event and the previous one in case of \textit{delay}. Time features are always range-normalized between $0$ and $1$, with the scope distinction differentiating if the normalization takes place before splitting events into pixels (\textit{absolute} feature) or after (\textit{relative} feature). In the case of \textit{ts}, \textit{absolute} means that the first and last events in the sequence have time feature $0$ and $1$, respectively, regardless of their position, whereas \textit{relative} means that the previous condition holds for each position $\xy$. Note that we only consider relative delays since it is only meaningful to compute them between events of the same pixel. Finally, we always add the polarity, obtaining a 2-value feature $f^{\xy}_i$. \textit{Delay relative} and \textit{ts absolute} are those providing the best results, with \textit{ts relative} having higher variance. We select \textit{delay relative} as the best configuration.
In Table~\ref{multitable:NCARS_Ev2Vid_Norm_Grads_EST_Bins}\subref*{table:NCARS_Ev2Vid_ResNet18_TimeEncodings_Norm_Grads} we also show the effect of applying the same frame normalization used while pre-training the ResNet backbone on ImageNet also to the Matrix-LSTM output. While performing normalization makes sense when training images are very similar to those used in pre-training, as in Events-to-Video~\cite{rebecq2019events}, we found out that in our case, where no constraint is imposed on the appearance of reconstructions, this does not improve the performance.

\begin{table}[t]
    \centering
    \caption{Results on N-Cars with ResNet18--EST: \textbf{(a)} \textit{polarity} + \textit{global ts} + \textit{local ts} encoding, optional SELayer and variable number of bins; \textbf{(b)} \textit{polarity} + \textit{global ts} + \textit{local ts} encoding, SELayer and variable number of channels}
    \label{multitable:NCARS_EST_SELayer_Channels_Bins}
    
    \subfloat[\label{table:NCARS_EST_ResNet18_SELayer_Bins}]{\begin{minipage}{.56\textwidth}
        \centering
        
        \resizebox{0.98\linewidth}{!}{%
        \begin{tabular}{c|ccccc}
        SE         & $2$ bins         & $4$ bins         & $9$ bins         & $16$ bins \\ \hline
                   & $\mathbf{93.46\pm0.84\%}$ & $92.68\pm0.62\%$ & $\mathbf{93.21\pm0,49\%}$ & $92.01\pm0.45\%$ \\
        \checkmark & $\mathbf{93.71\pm0.93\%}$ & $92.90\pm0.62\%$ & $\mathbf{93.30\pm0,47\%}$ & $92.44\pm0.43\%$
        \end{tabular}%
        }
    \end{minipage}}
    \subfloat[\label{table:NCARS_EST_ResNet18_SELayer_Channels}]{\begin{minipage}{.44\textwidth}
        \centering
        
        \resizebox{0.98\linewidth}{!}{%
        \begin{tabular}{c|ccc}
         & \multicolumn{3}{c}{Channels} \\
        bins  & 4                 & 8                 & 16 \\ \hline
        1       & $93.88\pm0.87\%$  & $93.60\pm0.30\%$  & $\mathbf{94.37\pm0.40\%}$ \\
        2       & $93.05\pm0.92\%$  & $93.97\pm0.52\%$  & $\mathbf{94.09\pm0.29\%}$ \\
        
        bins  & 4                 & 7                 & 8                 \\ \hline
        9       & $92.42\pm0.65\%$  & $93.56\pm0.46\%$  & $93.49\pm0.84\%$  
        \end{tabular}%
        }
    \end{minipage}}
\end{table}

\paragraph{\textbf{Matrix-LSTM + ResNet-EST.}} We continue the experiments on N-Cars by considering \textit{ResNet--EST} as baseline, where we explore the effect of using bins, i.e., intervals, on the quality of Matrix-LSTM surfaces. 
Since multiple intervals are involved, we distinguish between \textit{global} and \textit{local} temporal features. The first type is computed on the original sequence $\E$, before splitting events into intervals, whereas the latter locally, within the interval scope $\E_{\tau_b}$. For \text{local} features we consider the best options we identified on ResNet-Ev2Vid, namely \textit{delay relative} and \textit{ts absolute}, while we only consider \textit{ts} as global feature since a global delay loses meaning after interval splitting. Results are reported in Table~\ref{multitable:NCARS_Ev2Vid_Norm_Grads_EST_Bins}\subref*{table:NCARS_EST_ResNet18_TimeEncodings_Bins} where values for single bin are missing since there is no distinction between \textit{global} and \textit{local} scope. Adding a global feature consistently improves performance. This can indeed help the LSTM network in performing integration conditioned on a global timescale and thus enabling the extraction of temporal consistent features. We use \textit{global ts} + \textit{local ts} features in next experiments, since this provides better performance and reduced variance, and always add the polarity feature.

The next set of experiments was designed to select the optimal number of bins, searching for the best $B = 2, 4, 9, 16$ as done in EST, while using a fixed \textit{polarity} + \textit{global ts} + \textit{local ts} configuration. In these experiments, we also make use of the SELayer~\cite{hu2018squeeze}, a self-attention operation specifically designed to correlate channels. Being the number of channels limited, we always use a reduction factor of $1$. Please refer to the paper~\cite{hu2018squeeze} for more details. As reported in Table~\ref{multitable:NCARS_EST_SELayer_Channels_Bins}\subref*{table:NCARS_EST_ResNet18_SELayer_Bins}, adding the layer consistently improves performance. We explain this by noticing that surfaces computed on successive intervals are naturally correlated and, thus, explicitly modeling this behavior helps in extracting richer features. Finally, we perform the last set of experiments to select the Matrix-LSTM hidden size (which also controls the number of output channels). Results are reported in Table~\ref{multitable:NCARS_EST_SELayer_Channels_Bins}\subref*{table:NCARS_EST_ResNet18_SELayer_Channels}. Note that we only consider $4, 7, 8$ channels with $9$ bins to limit the total number of channels after concatenation.

\paragraph{\textbf{Matrix-LSTM vs. ConvLSTM.}}
\label{sec:convlstm_comparison}

\begin{figure}[t]
\centering
\subfloat[\label{tab:matrixVSconvLSTM}]{\begin{minipage}{.52\textwidth}
    \centering
    \resizebox{0.98\textwidth}{!}{%
    \begin{tabular}{cc|cc}
     &  & 3x3 & 5x5 \\ \hline
    \multirow{2}{*}{Matrix-LSTM} 
    & delay rel & $\mathbf{95.05 \pm 0.96\%}$ & $\mathbf{93.38 \pm 0.64\%}$ \\
     & ts abs & $\mathbf{94.92 \pm 0.74\%}$ & $\mathbf{94.34 \pm 0.94\%}$ \\ \hline
    \multicolumn{1}{l}{\multirow{2}{*}{ConvLSTM}} & delay rel & \multicolumn{1}{l}{$92.33 \pm 0.41\%$} & \multicolumn{1}{l}{$92.65 \pm 0.78\%$} \\
    \multicolumn{1}{l}{} & ts abs & \multicolumn{1}{l}{$93.97 \pm 1.30\%$} & \multicolumn{1}{l}{$93.61 \pm 1.59\%$}
    \end{tabular}%
    }
\end{minipage}}
\subfloat[\label{fig:matrixVSconv_space_time}]{\begin{minipage}{.43\textwidth}
    \centering
    \resizebox{0.88\textwidth} {!} {
    \input{data/matrixVSconv/plot_matrixVSconv}%
    }
\end{minipage}}

\caption{\textbf{(a)} Comparison between Matrix-LSTM and ConvLSTM on N-Cars. \textbf{(b)} Space and time relative improvements of Matrix-LSTM over ConvLSTM as a function of input density (from $10\%$ to $100\%$ with $30\%$ steps). Colors refer to different density, from low density (dark colors) to high density (light colors)}
\label{multifig:matrixVSconvLSTM_matrixVSconv_space_time}
\end{figure}
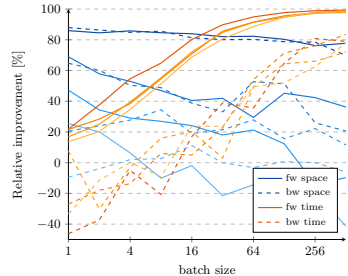

In Table \ref{multifig:matrixVSconvLSTM_matrixVSconv_space_time}\subref*{tab:matrixVSconvLSTM} we compare Matrix-LSTM with ConvLSTM \cite{shi2015convolutional} for different choices of kernel size on the N-Cars \cite{sironi2018hats} dataset using the Ev2Vid--ResNet18 backbone. When using ConvLSTM, events are densified in a volume $\tilde{\mathcal{E}}_{dense}$ of shape $N \times T_{max}^{(x,y)} \times H \times W \times F$. Matrix-LSTM performs better on all configurations, despite achieving worst performance than the $1 \times 1$ Matrix-LSTM best configuration in Table~\ref{multitable:NCARS_Ev2Vid_Norm_Grads_EST_Bins}\subref*{table:NCARS_Ev2Vid_ResNet18_TimeEncodings_Norm_Grads}. Event surfaces produced by the Matrix-LSTM layer are indeed more blurry with larger receptive fields and this may prevent the subsequent ResNet backbone from extracting effective features. Using a $1 \times 1$ kernel enables to focus on temporal information while the subsequent convolutional layers deal with spatial correlation.

ConvLSTM, instead, does not properly handle asynchronous data when large receptive fields are considered, and this may explains the performance difference with Matrix-LSTM. Indeed, since pixels at different locations most often fire at different times and with different frequencies, the $\tilde{\mathcal{E}}_{dense}[n, i, :, :, :]$ slice processed by the ConvLSTM in each iteration does not contain all simultaneous events. Using a large ConvLSTM receptive field means to compare a neighborhood of events occurred at different timestamps and therefore not necessarily correlated. Contrary to ConvLSTM, Matrix-LSTM allows for a greater flexibility when large receptive fields are considered since the original events arrival order is preserved and we do not require events to be densified during intermediate steps. We do not compare the two LSTMs on the $1 \times 1$ configuration since, when using $\tilde{\mathcal{E}}_{dense}$ as input to ConvLSTM, the two configurations compute the same transformation, despite ConvLSTM having to process more padded values. The two settings are indeed computationally equivalent only in the worst case in which all pixels in the batch happen to receive at least one event (i.e., $P = N \cdot H \cdot W$).

The $1 \times 1$ configurations are compared in terms of space and time efficiency in Figure \ref{multifig:matrixVSconvLSTM_matrixVSconv_space_time}\subref*{fig:matrixVSconv_space_time}. We use the two layers to extract a $224 \times 224$ frame from artificially generated events with increasing density, i.e., the ratio of pixels receiving at least one event. The reconstruction is performed using PyTorch \cite{steiner2019pytorch} on a $12$GB Titan Xp, by varying the batch size, the LSTM hidden size and the number of events in each active pixel (starting from $1$ and increasing by a factor of $2$ for the hidden size, while increasing by a factor of $10$ for the number of events, until allowed by GPU memory constraints). We compute the relative improvement of Matrix-LSTM in terms of sample reconstruction time and peak processing space (i.e., excluding model and input space) during both forward and backward passes, and finally aggregate the results by batch size computing the mean improvement over all the trials. Matrix-LSTM performs better than ConvLSTM on prediction time, with the time efficiency improving as the batch size increases, while worst than ConvLSTM on memory efficiency in very dense surfaces ($> 70\%$ density). However, this situation is quite uncommon in event-cameras since they only generate events when brightness changes are detected. Uniform parts of the scene that remain unchanged, despite the camera movement, do not appear in the event stream. For instance, the background sky and road in MVSEC \cite{zhu2018multivehicle} make \textit{outdoor\_day} sequences only have an average $10\%$ of active pixels.

\paragraph{\textbf{Discussion.}} Results of the top performing configurations for both \textit{ResNet-Ev2Vid} and \textit{ResNet-EST} variants on both N-Cars and N-Caltech101 are reported in Table~\ref{table:classification-sota}. We use \textit{relative delay} with \textit{ResNet-Ev2Vid} and \textit{global ts} + \textit{local ts} with \textit{ResNet-EST}.
Through an extensive evaluation, we show that using Matrix-LSTM representation as input to the baseline networks and training them jointly improves performance by a good margin. Indeed, using the ResNet34-Ev2Vid setup, our solution sets a new state-of-the-art on N-Cars, even surpassing the Events-to-Video model that was trained to extract realistic reconstructions. 
The same does not happen on N-Caltech101, whose performance usually greatly depends on pre-training also on the original image-based version, and where Events-to-Video has therefore advantage. Despite this, our model only performs $0.9\%$ worse than the baseline. On the ResNet-EST configuration, the model performs consistently better on N-Cars, while slightly worse on N-Caltech101 on most configurations. However, we remark that search for the best configuration was indeed performed on N-Cars, while a hyper-parameter search directly performed on N-Caltech101 would have probably lead to better results.

\begin{table}[t]
\caption{Matrix-LSTM best configurations compared to state-of-the-art}
\label{table:classification-sota}
\centering
\resizebox{0.665\linewidth}{!}{%
\begin{tabular}{c|cccc}
Method                                      & Classifier
                                            & \begin{tabular}[c]{@{}c@{}}Channels\\ (bins)\end{tabular}
                                                                                        & N-Cars   & N-Caltech101 \\ \hline \hline

H-First~\cite{orchard2015hfirst}            & spike-based           &  -                & $56.1$         & $0.54$ \\
HOTS~\cite{lagorce2016hots}                 & histogram similarity  &  -                & $62.4$         & $21.0$ \\
Gabor-SNN~\cite{sironi2018hats}             & SVM                   &  -                & $78.9$         & $19.6$ \\  \hline \hline
\multirow{3}{*}{HATS~\cite{sironi2018hats}} & SVM                   &  -                & $90.2$         & $64.2$ \\
                                            & ResNet34--EST~\cite{gehrig2019end} &  -   & $90.9$         & $69.1$ \\
                                        & ResNet18--Ev2Vid~\cite{rebecq2019events}& -   & $90.4$         & $70.0$ \\ \hline \hline

Ev2Vid~\cite{rebecq2019events}              & ResNet18--Ev2Vid      &  3        & $91.0$          & $\mathbf{86.6}$ \\ \hline
\multirow{2}{*}{\begin{tabular}[c]{@{}c@{}} \textbf{Matrix-LSTM} \\ \textbf{(Ours)}\end{tabular}}
                                            & ResNet18--Ev2Vid      &  3 (1)    & $\mathbf{95.80 \pm 0.53}$ & $84.12 \pm 0.84$ \\
                                            & ResNet34--Ev2Vid      &  3 (1)    & $\mathbf{95.65 \pm 0.46}$ & $85.72 \pm 0.37$ \\ \hline \hline

\multirow{2}{*}{EST~\cite{gehrig2019end}}   & ResNet34--EST         &  2 (9)    & $92.5$                    & $81.7$ \\
                                            & ResNet34--EST         &  2 (16)   & $92.3$                    & $83.7$ \\ \hline 
\multirow{6}{*}{\begin{tabular}[c]{@{}c@{}} \textbf{Matrix-LSTM} \\ \textbf{(Ours)}\end{tabular}}

                                            & ResNet18--EST         &  16 (1)   & $\mathbf{94.37 \pm 0.40}$ & $81.24 \pm 1.31$ \\ 
                                            & ResNet34--EST         &  16 (1)   & $\mathbf{94.31 \pm 0.43}$ & $78.98 \pm 0.54$ \\ \cline{2-5}
                                            & ResNet18--EST         &  16 (2)   & $\mathbf{94.09 \pm 0.29}$ & $83.42 \pm 0.80$ \\ 
                                            & ResNet34--EST         &  16 (2)   & $\mathbf{94.31 \pm 0.44}$ & $80.45 \pm 0.55$ \\ \cline{2-5}
                                            & ResNet18--EST         &  2 (16)   & $\mathbf{92.58 \pm 0.68}$ & $\mathbf{84.31 \pm 0.59}$ \\
                                            & ResNet34--EST         &  2 (16)   & $92.15 \pm 0.73$          & $83.50 \pm 1.24$ \\  \hline \hline
\end{tabular}%
}
\end{table}

\subsection{Optical flow prediction}

\label{sec:opticalflow}
For the evaluation of optical flow prediction we used the MVSEC~\cite{zhu2018multivehicle} suite. Fusing event-data with lidar, IMU, motion capture and GPS sources, MVSEC is the first event-based dataset to provide a solid benchmark in real urban conditions. The dataset provides ground truth information for depth and vehicle pose and was later extended in \cite{Zhu2018ev-flownet} with optical flow information extracted from depth-maps. The dataset has been recorded on a range of different vehicles and features both indoor and outdoor scenarios and different lighting conditions.

\paragraph{\textbf{Network Architecture.}}
\label{par:opticalflow-network}
We used the EV-FlowNet~\cite{Zhu2018ev-flownet} architecture as reference model. To perform a fair comparison between Matrix-LSTM and the original hand-crafted features, we built our model on top of its publicly available codebase~\cite{daniilidisgroup2019github}. The code contains few minor upgrades over the paper version, which we made sure to remove as reported in the supplementary materials.


The original network uses a $4$-channels event-surface, collecting in pairs of separate channels based on the event polarity, the timestamp of the most recent event, and the number of events occurred in every spatial location. We replaced this representation with a Matrix-LSTM making use of $4$ output channels, as well.
We trained the model on the \textit{outdoor\_day1} and \textit{outdoor\_day2} sequences for $300,000$ iterations, as in the original paper. We used the ADAM optimizer with batch size $8$, and an initial learning rate of $10^{-5}$, exponentially decayed every $4$ epochs by a factor of $0.8$. We noticed that EV-FlowNet is quite unstable at higher learning rates, while Matrix-LSTM could benefit from larger rates, so we multiply its learning rate, i.e., the Matrix-LSTM gradients, by a factor of $10$ during training. Test was performed on a separate set of recordings, namely \textit{indoor\_flying1}, \textit{indoor\_flying2} and \textit{indoor\_flying3}, which are visually different from the training data. The network performance is measured in terms of average endpoint error (AEE), defined as the distance between the endpoints of the predicted and ground truth flow vectors. In addition, as proposed in the KITTI benchmark~\cite{Menze2015CVPR} and as done in~\cite{Zhu2018ev-flownet}, we report the percentage of outliers, namely points with endpoint error greater than $3$ pixels and $5\%$ of the magnitude ground truth vector. Finally, following the procedure used in~\cite{Zhu2018ev-flownet}, we only report the error computed in spatial locations where at least one event was generated. 

\begin{table*}[t]\centering
\caption{Optical flow estimation on MVSEC}
\label{tab:opticalflow}

\centering
\resizebox{0.76\textwidth}{!}{%
\begin{tabular}{c|c|cc|cc|cc}
\multirow{2}{*}{Method} & \multirow{2}{*}{} & \multicolumn{2}{c|}{\textit{indoor\_flying1}} & \multicolumn{2}{c|}{\textit{indoor\_flying2}} & \multicolumn{2}{c}{\textit{indoor\_flying3}} \\
                                            &                   & AEE    & $\%$Outlier & AEE & $\%$Outlier & AEE & $\%$Outlier \\ \hline \hline
Two-Channel Image~\cite{maqueda2018event}   &                   & $1.21$   & $4.49$  & $2.03$  & $22.8$  & $1.84$  & $17.7$ \\
EV-FlowNet~\cite{Zhu2018ev-flownet}         &                   & $1.03$   & $2.20$  & $1.72$  & $15.1$  & $1.53$  & $11.9$ \\ \hline
Voxel Grid~\cite{zhu2019unsupervised}       &                   & $0.96$   & $1.47$  & $1.65$  & $14.6$  & $1.45$  & $11.4$ \\ \hline
\multirow{2}{*}{EST~\cite{gehrig2019end}}   & exp. kernel       & $0.96$   & $1.27$  & $1.58$  & $10.5$  & $1.40$  & $9.44$ \\
                                            & learnt kernel     & $0.97$   & $0.91$  & $1.38$  & $8.20$  & $1.43$  & $6.47$ \\ \hline \hline
\multirow{5}{*}{\begin{tabular}[c]{@{}c@{}}Matrix-LSTM\\ (Ours)\end{tabular}} 
                                            & 1 bin             & $1.017$  & $2.071$ & $1.642$ & $13.88$ & $1.432$ & $10.44$ \\
                                            & 2 bins            & $\mathbf{0.829}$  & $\mathbf{0.471}$ & $\mathbf{1.194}$ & $\mathbf{5.341}$ & $\mathbf{1.083}$ & $\mathbf{4.390}$ \\
                                            & 4 bins            & $0.969$  & $1.781$ & $1.505$ & $11.63$ & $1.507$ & $12.97$ \\
                                            & 8 bins            & $0.881$  & $0.672$ & $1.292$ & $6.594$ & $1.181$ & $5.389$ \\ \cline{2-8}
                                            & 2 bins + SELayer  & $\mathbf{0.821}$  & $\mathbf{0.534}$ & $\mathbf{1.191}$ & $\mathbf{5.590}$ & $\mathbf{1.077}$ & $\mathbf{4.805}$ \\ \hline \hline
\end{tabular}%
}

\end{table*}

\paragraph{\textbf{Results.}}
\label{par:opticalflow-results}
In the previous classification experiments, we observed that the type of temporal features and the number of bins play an important role in extracting effective representations. We expect time resolution to be a key factor of performance in optical flow, hence, we focus here on measuring how different interval choices impact on the flow prediction. 
We decided to always use the \textit{polarity} + \textit{global ts} + \textit{local ts} configuration, which worked well on N-Cars while considering different bin setups. Results are reported in Table~\ref{tab:opticalflow}. 

As performed on classification, we study the effect of adding a SELayer on the best performing configuration. Correlating the intervals slightly improves the AEE metric in all test sequences but increases the number of outliers. As expected, varying the number of bins has a great impact on performance. The AEE metric, indeed, greatly reduces by simply considering two intervals instead of one. Interestingly, we achieved the best performance by considering only $2$ intervals, as adding more bins hurts performance. We believe this behavior resides on the nature of optical flow prediction, where the network is implicitly asked to compare two distinct temporal instants. This configuration consistently improves the baseline up to $30.76\%$ on \textit{indoor\_flying2}, highlighting the capability of the Matrix-LSTM to adapt also to low-level tasks.

\subsection{Time performance analysis}
\label{sec:time-results}

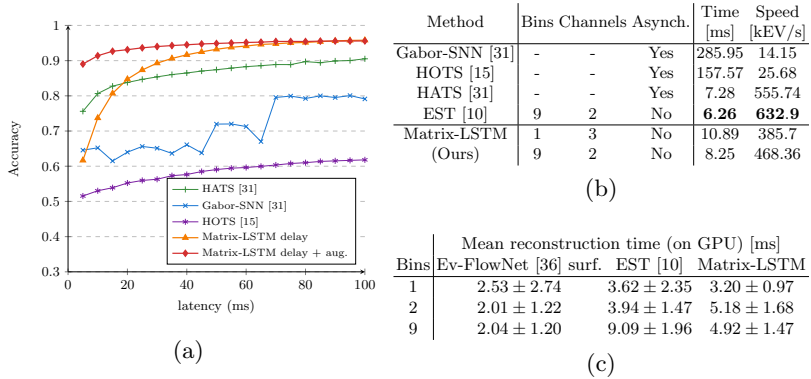
\begin{figure}[t]
\centering
   \begin{minipage}{.42\textwidth} 
       \subfloat[\label{fig:latency}]{\begin{minipage}{\linewidth}
       \centering
            \input{data/plot_latency.tex}
        \end{minipage}}
    \end{minipage} 
    \begin{minipage}{.462\textwidth} 
        \subfloat[\label{tab:timings}]{\begin{minipage}{\linewidth}
            \centering
            \resizebox{0.98\linewidth}{!}{%
            \begin{tabular}{c|ccc|cc}
            Method                          & Bins  & Channels  & Asynch.  &    \begin{tabular}[c]{@{}c@{}}Time\\ $[$ms$]$\end{tabular} & 
                                                                             \begin{tabular}[c]{@{}c@{}}Speed\\ $[$kEV/s$]$\end{tabular} \\ \hline
            Gabor-SNN~\cite{sironi2018hats} & -     & -         & Yes           & $285.95$            & $14.15$ \\
            HOTS~\cite{lagorce2016hots}     & -     & -         & Yes           & $157.57$            & $25.68$ \\
            HATS~\cite{sironi2018hats}      & -     & -         & Yes           & $7.28$              & $555.74$ \\
            EST~\cite{gehrig2019end}        & $9$   & $2$       & No            & $\mathbf{6.26}$   & $\mathbf{632.9}$ \\ \hline
            \multirow{2}{*}{\begin{tabular}[c]{@{}c@{}}Matrix-LSTM\\ (Ours)\end{tabular}} 
                                            & $1$   & $3$         & No            & $10.89$             & $385.7$ \\
                                            & $9$   & $2$         & No            & $8.25$              & $468.36$
            \end{tabular}%
            }
        \end{minipage}}
        \vfill
        \subfloat[\label{tab:timings-mvsec}]{\begin{minipage}{\linewidth}
            \centering
            \resizebox{0.98\linewidth}{!}{%
            \begin{tabular}{c|ccc}
            \textbf{} & \multicolumn{3}{c}{Mean reconstruction time (on GPU) [ms]}       \\
            Bins        & Ev-FlowNet~\cite{Zhu2018ev-flownet} surf. 
                        & EST~\cite{gehrig2019end}  
                        & Matrix-LSTM \\ \hline
            $1$         & $2.53 \pm 2.74$      & $3.62 \pm 2.35$    & $3.20 \pm 0.97$ \\
            $2$         & $2.01 \pm 1.22$      & $3.94 \pm 1.47$    & $5.18 \pm 1.68$ \\
            $9$         & $2.04 \pm 1.20$      & $9.09 \pm 1.96$    & $4.92 \pm 1.47$
            \end{tabular}%
            }
        \end{minipage}}
    \end{minipage}
    
    \caption{\textbf{(a)} Accuracy as a function of latency (adapted from \cite{sironi2018hats}).
             \textbf{(b)} Average sample computation time on N-Cars and number of events processed per second.
             \textbf{(c)} Average time to reconstruct the event surface in MVSEC test sequences}
    \label{multifig:latency_timings}
\end{figure}

We compared the time performance of Matrix-LSTM with other event representations following EST~\cite{gehrig2019end} and HATS~\cite{sironi2018hats} evaluation procedure. In Table~\ref{multifig:latency_timings}\subref*{tab:timings} we report the time required to compute features on a sample averaged over the whole N-Cars training dataset for both ResNet--Ev2Vid and ResNet--EST configurations. Our surface achieves similar time performance than both HATS and EST, performing only ${\sim}2$ms slower than EST on the same setting ($9$ bins and $2$ channels). Similarly, in Table \ref{multifig:latency_timings}\subref*{tab:timings-mvsec}, we compute the mean surface reconstruct time for MVSEC \textit{indoor\_flying} test sequences.
While EST can exploit parallel batch computation of events within the same sample, since each event feature is processed independently, Matrix-LSTM relies on sequential computation to reconstruct the surface. The custom CUDA kernels we designed, 
however, enable bins and pixel sequences to be processed in parallel, drastically reducing the processing time. Please, refer to the additional materials for more details. All evaluations are performed with PyTorch on a GeForce GTX 1080Ti GPU.

In Figure \ref{multifig:latency_timings}\subref*{fig:latency} we analyze the accuracy-vs-latency trade-off on the N-Cars dataset, as proposed in \cite{sironi2018hats}, using the ResNet18-Ev2Vid configuration. While the performance of the model, trained on $100$ms sequences, significantly drops when very few milliseconds of events are considered, the proposed method still shows good generalization, achieving better performance than the baselines when more than $20$ms of events are used. However, fixing the performance loss on small latencies is just a matter of training augmentation: by randomly cropping sequences to variable lengths (from $5$ms to $100$ms), our method consistently improves the baselines, dynamically adapting to sequences of different lengths.

\section{Conclusion}
\label{sec:conclusion}

We proposed Matrix-LSTM, an effective method for learning dense event representations from event-based data. By modeling the reconstruction with a spatially shared LSTM we obtained a fully differentiable procedure that can be trained end-to-end to extract the event representation that best fits the task at hand. Focusing on efficiently handling asynchronous data, Matrix-LSTM preserves sparsity during computation and surpasses other popular LSTM variants on space and time efficiency when processing sparse inputs. In this regard, we proposed an efficient implementation of the method that exploits parallel batch-wise computation and demonstrated the effectiveness of the Matrix-LSTM layer on multiple tasks, improving the state-of-the-art of object classification on N-Cars by $3.3\%$ and the performance on optical flow prediction on MVSEC by up to $23.07\%$ over previous differentiable techniques~\cite{gehrig2019end}. Although we only integrate windows of events, the proposed mechanism can be extended to process a continuous streams thanks to the LSTM memory that is able to update its representation as soon as a new event arrives. As a future line of research, we plan to explore the use of Matrix-LSTM for more complex tasks such as gray-scale frame reconstruction \cite{rebecq2019events}, ego-motion and depth estimation \cite{zhu2019unsupervised,ye2018unsupervised}.

\paragraph{\textbf{Acknowledgments.}}
We thank Alex Zihao Zhu for his help on replicating Ev-FlowNet results and the ISPL group at Politecnico di Milano for GPU support. This research is supported from project TEINVEIN, CUP: E96D17000110009 - Call "Accordi per la Ricerca e l'Innovazione", cofunded by POR FESR 2014-2020 (Regional Operational Programme, European Regional Development Fund).

\appendix
\appendixtitleon
\appendixtitletocon
\begin{appendices}
\renewcommand{\thesection}{Appendix \Alph{section}} 
\renewcommand{\thesubsection}{\Alph{section}.\arabic{subsection}}
\input{sup_content.tex}
\end{appendices}

\clearpage
%
%
\bibliographystyle{splncs04}
\bibliography{egbib}
\end{document}

%% file: data/matrixVSconv/plot_matrixVSconv.tex
  \begin{tikzpicture}
  \begin{axis}[
      enlargelimits=false,
      axis y line=left,
      axis x line=bottom,
      yticklabel pos=left,
      legend pos=south east,
      ylabel={Relative improvement $[\%]$},
      xlabel={batch size},
      ymin=-50, ymax=100,
      symbolic x coords={1,2,4,8,16,32,64,128,256,512},
      ytick={-40,-20,0,20,40,60,80,100},
      ymajorgrids=true,
      grid style=dashed,
      legend style={cells={align=left},
                    font=\scriptsize},
      legend cell align={left},
  ]
  \addplot[thick, color=MaterialBlue900]
      table[x index=0,y index=1,col sep=comma]
      {data/matrixVSconv/10density_fw_space.txt};
  \addplot[thick, forget plot, color=MaterialBlue700]
      table[x index=0,y index=1,col sep=comma]
      {data/matrixVSconv/40density_fw_space.txt};
  \addplot[thick, forget plot, color=MaterialBlue500]
      table[x index=0,y index=1,col sep=comma]
      {data/matrixVSconv/70density_fw_space.txt};
  \addplot[thick, forget plot, color=MaterialBlue300]
      table[x index=0,y index=1,col sep=comma]
      {data/matrixVSconv/100density_fw_space.txt};
      
  \addplot[thick, dashed, color=MaterialBlue900]
      table[x index=0,y index=1,col sep=comma]
      {data/matrixVSconv/10density_bw_space.txt};
  \addplot[thick, dashed, forget plot, color=MaterialBlue700]
      table[x index=0,y index=1,col sep=comma]
      {data/matrixVSconv/40density_bw_space.txt};
  \addplot[thick, dashed, forget plot, color=MaterialBlue500]
      table[x index=0,y index=1,col sep=comma]
      {data/matrixVSconv/70density_bw_space.txt};
  \addplot[thick, dashed, forget plot, color=MaterialBlue300]
      table[x index=0,y index=1,col sep=comma]
      {data/matrixVSconv/100density_bw_space.txt};
  
  \addplot[thick, color=MaterialOrange900]
      table[x index=0,y index=1,col sep=comma]
      {data/matrixVSconv/10density_fw_time.txt};
  \addplot[thick, forget plot, color=MaterialOrange700]
      table[x index=0,y index=1,col sep=comma]
      {data/matrixVSconv/40density_fw_time.txt};
  \addplot[thick, forget plot, color=MaterialOrange500]
      table[x index=0,y index=1,col sep=comma]
      {data/matrixVSconv/70density_fw_time.txt};
  \addplot[thick, forget plot, color=MaterialOrange300]
      table[x index=0,y index=1,col sep=comma]
      {data/matrixVSconv/100density_fw_time.txt};
      
  \addplot[thick, dashed, color=MaterialOrange900]
      table[x index=0,y index=1,col sep=comma]
      {data/matrixVSconv/10density_bw_time.txt};
  \addplot[thick, dashed, forget plot, color=MaterialOrange700]
      table[x index=0,y index=1,col sep=comma]
      {data/matrixVSconv/40density_bw_time.txt};
  \addplot[thick, dashed, forget plot, color=MaterialOrange500]
      table[x index=0,y index=1,col sep=comma]
      {data/matrixVSconv/70density_bw_time.txt};
  \addplot[thick, dashed, forget plot, color=MaterialOrange300]
      table[x index=0,y index=1,col sep=comma]
      {data/matrixVSconv/100density_bw_time.txt};
      
  \addlegendentry{fw space};
  \addlegendentry{bw space};
  \addlegendentry{fw time};
  \addlegendentry{bw time};
 
  \end{axis}
  \end{tikzpicture}

%% file: data/plot_latency.tex
\begin{tabular}{c}
\centering
\resizebox{\textwidth} {!} {
  \begin{tikzpicture}
  \begin{axis}[
      enlargelimits=false,
      axis y line=left,
      axis x line=bottom,
      yticklabel pos=left,
      legend pos=south east,
      xlabel={latency (ms)},
      ylabel={Accuracy},
      xmin=0, xmax=100,
      ymin=0.3, ymax=1.0,
      symbolic x coords={0,5,10,15,20,25,30,35,40,45,50,55,60,65,70,75,80,85,90,95,100},
      xtick={0, 20, 40, 60, 80, 100},
      ytick={0.2,0.3,0.4,0.5,0.6,0.7,0.8,0.9,1.0},
      ymajorgrids=true,
      grid style=dashed,
      legend style={font=\scriptsize},
      legend cell align={left},
      legend entries={HATS \cite{sironi2018hats},
                      Gabor-SNN \cite{sironi2018hats},
                      HOTS \cite{lagorce2016hots},
                      Matrix-LSTM delay,
                      Matrix-LSTM delay + aug.}
  ]
  \addplot[
      color=MaterialGreen700,
      mark=+]
      table[x index=0,y index=1,col sep=comma]
      {data/latency_hats.txt};
  \addplot[
      color=MaterialBlue700,
      mark=x]
      table[x index=0,y index=1,col sep=comma]
      {data/latency_gaborsnn.txt};
  \addplot[
      color=MaterialPurple700,
      mark=asterisk]
      table[x index=0,y index=1,col sep=comma]
      {data/latency_hots.txt};
  \addplot[
      thick,
      color=MaterialOrange700,
      mark=triangle*]
      table[x index=0,y index=1,col sep=comma]
      {data/latency_delayRel.txt};
  \addplot[
      thick,
      color=MaterialRed700,
      mark=diamond*]
      table[x index=0,y index=1,col sep=comma]
      {data/latency_delayRel_Aug.txt};
  \end{axis}
  \end{tikzpicture}
}
\end{tabular}

%% file: sup_content.tex
\begin{figure}[t]
        \centering
        \includegraphics[width=0.6\linewidth]{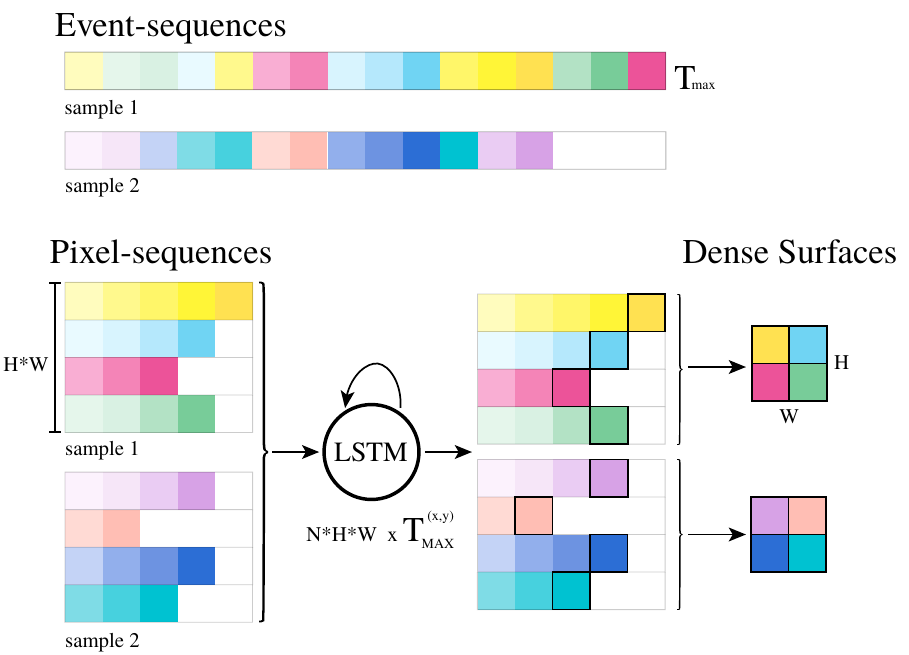}
        \vspace{-4pt}
        \caption{An example of the \emph{groupByPixel} operation on a batch of $N=2$ samples and a $2\times2$ pixel resolution. Different colors refer to different pixel locations while intensity indicates time. For clarity, the features dimension is not shown in the figure}
        \label{fig:groupbypixel}
\end{figure}

\section{Implementation}

The Matrix-LSTM feature extraction process can be implemented efficiently by means of two order-aware reshape operations. These two operations, namely \textit{groupByPixel} and \textit{groupByTime}, allow event streams to be split based on the pixel location and temporal bin. After being reshaped, the input is ready to be processed by a single LSTM network, implementing parameter sharing across all pixel locations and temporal windows. In the following we give a detailed overview of the two reshape operators.

\subsection{GroupByPixel}
\label{sec:groupbypixel}

This operation translates from event-sequences to pixel-sequences. Let $X$ be a tensor of shape $N \times T_{max} \times F$, representing the features $f^{\xy}_{n,i}$ of a batch of $N$ samples, where $T_{max}$ is the length of the longest sequence in the batch. We define the \textit{groupByPixel} mapping on $X$ as an order-aware reshape operation that rearranges the events into a tensor of pixel-sequences of shape $P \times T^{\xy}_{max} \times F$ where $T^{\xy}_{max}$ is the length of the longest pixel sequence $\E^{\xy}_n$ and $P$ is the number of active pixels (i.e.,  having at least one event) in the batch, which equals $N \cdot H \cdot W$ only in the worst case. Pixel-sequences shorter than $T^{\xy}_{max}$ are padded with zero events to be processed in parallel.

The tensor thus obtained is then processed by the LSTM cell that treats samples in the first dimension independently, effectively implementing parameter sharing and applying the transformation in parallel over all the pixels. The LSTM output tensor, which has the same shape of the input one, is then sampled by taking the output corresponding to the last event in each pixel-sequence $\E^{\xy}_n$, ignoring values computed on padded values, and the obtained values are then used to populate the dense representation. To improve efficiency, for each pixel-sequence $\E^{\xy}_n$, \textit{groupByPixel} keeps also track of the original spatial position $\xy$, the index of the sample inside the batch and the length of the pixel-sequence $T^{\xy}_n$, namely the index of the last event before padding. Given this set of indexes, the densification step can be performed as a simple slicing operation. See Figure~\ref{fig:groupbypixel} for visual clues. \textit{groupByPixel} is implemented as a custom CUDA kernel that processes each sample in parallel and places each event feature in the output tensor maintaining the original temporal order. 

\subsection{GroupByTime}
The Matrix-LSTM variant that operates on temporal bins performs a similar pre-processing step. Each sample in the batch is divided into a fixed set of intervals.
The \textit{groupByTime} cuda kernel pre-processes the input events generating a $N * B \times T_{max}^b \times F$ tensor where the $B$ bins are grouped in the first dimension and taking care of properly padding intervals ($T_{max}^b$ is the length of the longest bin in the batch). The Matrix-LSTM mechanism is then applied as usual and the resulting ${N*B \times H \times W \times C}$ tensor is finally reshaped into a ${N \times H \times W \times B*C}$ event-surface.

\section{Matrix-LSTM vs ConvLSTM}

\begin{table}[t]
\caption{Comparison between Matrix-LSTM and ConvLSTM on both Ev2Vid and EST ResNet18 configurations on the N-Cars dataset}
\label{tab:extended_MatrixVsConvLSTM}

\centering
\resizebox{0.85\textwidth}{!}{%
\begin{tabular}{cc|cc|cc}
 &  & \multicolumn{2}{c|}{delay relative} & \multicolumn{2}{c}{ts absolute} \\
 &  & $3 \times 3$ & $5 \times 5$ & $3 \times 3$ & $5 \times 5$ \\ \hline
\multicolumn{1}{c|}{\multirow{2}{*}{\begin{tabular}[c]{@{}c@{}}Ev2Vid with\\ 3 chans, 1 bin\end{tabular}}} & Matrix-LSTM (ours) & $\mathbf{95.05 \pm 0.96\%}$ & $\mathbf{93.38 \pm 0.64\%}$ & $\mathbf{94.92 \pm 0.74\%}$ & $\mathbf{94.34 \pm 0.94\%}$ \\
\multicolumn{1}{c|}{} & ConvLSTM \cite{shi2015convolutional} & $92.33 \pm 0.41\%$ & $92.65 \pm 0.78\%$ & $93.97 \pm 1.30\%$ & $93.61 \pm 1.59\%$ \\ \hline
\multicolumn{1}{c|}{\multirow{2}{*}{\begin{tabular}[c]{@{}c@{}}EST with\\ 16 chans, 1 bin\end{tabular}}} & Matrix-LSTM (ours) & $\mathbf{93.14 \pm 0.77\%}$ & $\mathbf{92.18 \pm 0.28\%}$ & $\mathbf{92.83 \pm 1.32\%}$ & $\mathbf{92.15 \pm 0.67\%}$ \\
\multicolumn{1}{c|}{} & ConvLSTM \cite{shi2015convolutional} & $ 90.39\pm 0.94\%$ & $90.73 \pm 1.05\%$ & $92.52 \pm 1.26\%$ & $92.05 \pm 0.56\%$
\end{tabular}%
}
\end{table}

In Figure 2a of the paper we report a comparison between ConvLSTM and Matrix-LSTM using the Ev2Vid--ResNet18 configuration, i.e., the configuration of choice for all comparisons in the paper. For completeness, in Table \ref{tab:extended_MatrixVsConvLSTM} we extend the evaluation also to EST, using the $16$ channels and $1$ bin setting, which is one of our best performing EST configuration on N-Cars. Matrix-LSTM performs better on all experiments, highlighting its capabilities to better handle asynchronous event-based data if compared to ConvLSTM. We also highlight that the performance improvement is greater on \textit{delay relative} temporal features than on \textit{ts absolute} ones. ConvLSTM, indeed, processes temporal slices containing potentially uncorrelated events that happened at different time instants. While delays are always consistent within each pixel sequence, they are not within the ConvLSTM kernel receptive field. Using an absolute temporal encoding alleviates this issue on both Ev2Vid and EST architectures, while still performing worst than Matrix-LSTM. Our extraction layer, indeed, preserves the original events arrival order within each receptive field during features extraction, which allows to achieve better performance both on \textit{ts absolute} and \textit{delay relative} input features. Moreover, structured \textit{delay relative} features perform better on Matrix-LSTM than simple absolute features.

\section{Time Performance}
\label{sec:time_performance}

\begin{figure*}[t]
\centering
    \subfloat[\label{fig:chans_vs_batch}]{\begin{minipage}{.49\textwidth}
        \centering
        \resizebox{\textwidth}{!}{%
            \input{data/plot_chan_vs_batch_time.tex}
        }
        \vspace{-8pt}
    \end{minipage}}
    \hspace{4pt}
    \subfloat[\label{fig:bin_vs_batch}]{\begin{minipage}{.49\textwidth}
        \centering
        \resizebox{\textwidth}{!}{%
            \input{data/plot_bin_vs_batch_time.tex}
        }
    \end{minipage}}
    
    \caption{Number of processed events per second (dashed lines) and timing (solid lines) with varying number of channels \textbf{(a)}, and bins \textbf{(b)}}
    \label{multifig:chans_bins_vs_batch}
\end{figure*}
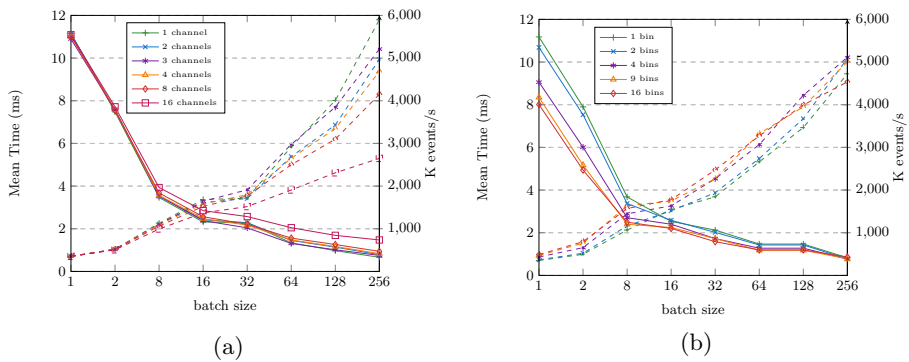

While the performance reported in Figure 3b of the paper are computed on each sample independently to enable a fair comparison with the other methods, in Figure \ref{multifig:chans_bins_vs_batch}\subref*{fig:chans_vs_batch} and Figure \ref{multifig:chans_bins_vs_batch}\subref*{fig:bin_vs_batch} we study instead how the mean time required to process a sample over all the N-Cars training dataset and the corresponding events throughput change as a function of the batch size. Both performance dramatically increase when multiple samples are processed simultaneously in batch. This is crucial at training time, when optimization techniques greatly benefit from batch computation. 

Furthermore, while increasing the number of output channels, for the same choice of batch size, increases the time required to process each sample (since the resulting Matrix-LSTM operates on a larger hidden state), increasing the number of bins has an opposite behaviour. Multiple intervals are indeed processed independently and in parallel by the Matrix-LSTM that has to process a smaller number of events in each spatial location, sequentially. In both configurations, finally, increasing the batch size reduces the mean processing time.

\section{Classification}

\begin{table}[]
\centering

\caption{Classification accuracy (\%) on the N-MNIST \cite{orchard2015converting} dataset.}
\label{tab:nmnist}

\resizebox{0.6\textwidth}{!}{%
\begin{tabular}{c|ccc}
Method & Classifier  & Channels (bins) & Accuracy        \\ \hline\hline
H-First
& spike-based            & -               &                 \\
HOTS \cite{lagorce2016hots}
& histogram similarity   & -               & $80.8$          \\
HATS \cite{sironi2018hats}
& SVM                    & -               & $99.1$          \\
G-CNN \cite{bi2019graph}
& Graph CNN              & -               & $98.5$          \\
RG-CNN \cite{bi2019graph}
& Graph CNN              & -               & $\mathbf{99.0}$          \\ \hline\hline
Events Count \cite{bi2019graph}
& ResNet50  & 2 (1)     & $98.4$           \\
Ev2Vid \cite{rebecq2019events}
& \begin{tabular}[c]{@{}c@{}}Ev2Vid custom\\ convnet\end{tabular} & 1 (1)               & $98.3$          \\ \hline
\textbf{\begin{tabular}[c]{@{}c@{}}Matrix-LSTM\\ (Ours)\end{tabular}} 
& \begin{tabular}[c]{@{}c@{}}Ev2Vid custom\\ convnet\end{tabular} & 1 (1)               & $\mathbf{98.9 \pm 0.21}$
\end{tabular}%
}
\end{table}

\begin{table}[]
\centering

\caption{Classification accuracy (\%) on the ASL-DVS \cite{bi2019graph} dataset.}
\label{tab:nmnist}

\resizebox{0.55\textwidth}{!}{%
\begin{tabular}{c|ccc}
Method & Classifier & Channels (bins) & Accuracy         \\ \hline\hline
G-CNN \cite{bi2019graph}
& Graph CNN  & -               & $87.5$           \\
RG-CNN \cite{bi2019graph}
& Graph CNN  & -               & $90.1$           \\ \hline\hline
Events Count \cite{bi2019graph}
& ResNet50 & 2 (1)           & $88.6$           \\
EST \cite{gehrig2019end}
& ResNet50  & 2 (1)           & $99.57$          \\ \hline
\textbf{\begin{tabular}[c]{@{}c@{}}Matrix-LSTM\\ (Ours)\end{tabular}}
& ResNet50  & 2 (1)           & $\mathbf{99.73 \pm 0.04}$
\end{tabular}%
}
\end{table}

We perform additional experiments on the N-MNIST dataset \cite{orchard2015converting} and on the newly introduced ASL-DVS \cite{bi2019graph} dataset. On N-MNIST we directly compare with the Ev2Vid \cite{rebecq2019events} reconstruction procedure, where the custom convolutional network proposed in \cite{rebecq2019events} is used as backbone, while we compare with the EST \cite{gehrig2019end} surface on ASL-DVS, making use of ResNet50 \cite{he2016deep} as backbone. On both cases, Matrix-LSTM performs better than other event-surface mechanisms and also outperforms alternative classification architectures.

\section{Optical Flow Prediction}

\subsection{Ev-FlowNet Baseline Results}
We performed optical flow experiments starting from the publicly available Ev-FlowNet codebase \cite{daniilidisgroup2019github} and replacing the original hand-crafted features with the proposed Matrix-LSTM layer. We first made sure to revert the baseline architecture to the original configuration, checking that we were able to replicate the paper results. Indeed, the public code contains minor upgrades over the paper version. We contacted the authors that provided us with the needed modifications. These consist of removing the batch normalization layers, setting to $2$ the number of output channels of the layer preceding the optical flow prediction layer, and disabling random rotations during training. For completeness, we report the results we obtained by training the baseline from scratch with these fixes in Table \ref{tab:opticalflow_SELayer}.

To test how the network adapts to different flow magnitudes, the Ev-FlowNet \cite{Zhu2018ev-flownet} was tested on two evaluation settings for each test sequence: with input frames and corresponding events that are one frame apart (denoted as \textit{dt=1}), and with frames and events four frames apart (denoted as \textit{dt=4}). While we were able to closely replicate the results of the first configuration (\textit{dt=1}), with a minor improvement in the \textit{indoor\_flying2} sequence, the performance we obtain on the \textit{dt=4} setup is instead worse on all sequences, as reported on the first two rows of Table \ref{tab:opticalflow_SELayer}.

Despite this discrepancy, which prevents the Matrix-LSTM performance on \textit{dt=4} settings to be directly compared with the results reported on the Ev-FlowNet paper, we can still evaluate the benefits of our surface on larger flow magnitudes. Indeed, this work evaluates the Matrix-LSTM layer based on the relative performance improvement obtained by substituting the original features with our layer. Using our Ev-FlowNet results as baseline, we show that Matrix-LSTM is able to improve the optical flow quality even on the \textit{dt=4} setting, highlighting the capability of the layer to adapt to different sequence lengths and movement conditions. We report an improvement of up to $30.426\%$ on \textit{dt=1} settings and up to $39.546\%$ on \textit{dt=4} settings using our results as baseline.

\begin{table*}[t]\centering
\caption{Effect of adding a Squeeze-and-Excitation layer on the optical flow prediction task}
\label{tab:opticalflow_SELayer}

\resizebox{\textwidth}{!}{%
\begin{tabular}{c|c|cc|cc|cc|cc|cc|cc}
\multirow{3}{*}{Method} & \multirow{2}{*}{} & \multicolumn{4}{c|}{\textit{indoor\_flying1}} & \multicolumn{4}{c|}{\textit{indoor\_flying2}} & \multicolumn{4}{c}{\textit{indoor\_flying3}} \\
&                   & \multicolumn{2}{c|}{\textit{dt=1}}      & \multicolumn{2}{c|}{\textit{dt=4}}        & \multicolumn{2}{c|}{\textit{dt=1}}      & \multicolumn{2}{c|}{\textit{dt=4}}        & \multicolumn{2}{c|}{\textit{dt=1}}      & \multicolumn{2}{c}{\textit{dt=4}}    \\
&                   & AEE               & $\%$Outlier        & AEE               & $\%$Outlier          & AEE               & $\%$Outlier       & AEE               & $\%$Outlier           & AEE              & $\%$Outlier        & AEE               & $\%$Outlier       \\ \hline

Ev-FlowNet \cite{Zhu2018ev-flownet}
& -                 & $1.03$            & $2.2$              & $2.25$            & $24.7$               & $2.12$            & $15.1$            & $4.05$            & $45.3$                & $1.53$           & $11.9$             & $3.45$            & $39.7$            \\
Ev-FlowNet (ours)
& -                 & $1.015$           & $2.736$            & $3.432$           & $48.685$             & $1.606$           & $12.089$          & $5.957$           & $63.226$              & $1.548$          & $11.937$           & $5.247$           & $57.662$            \\ \hline

\multirow{7}{*}{\begin{tabular}[c]{@{}c@{}}Matrix-LSTM\\ (Ours)\end{tabular}} 

& 1 bin             & $1.017$           & $2.071$            & $3.366$           & $42.022$             & $1.642$           & $13.89$           & $5.870$           & $65.379$              & $1.432$           & $10.44$           & $5.015$           & $57.094$          \\ \cline{2-14}
& 2 bins            & $0.829$           & $\mathbf{0.471}$   & $\mathbf{2.269}$  & $\mathbf{23.558}$    & $1.194$           & $\mathbf{5.341}$  & $\mathbf{3.946}$  & $\mathbf{42.450}$     & $1.083$           & $\mathbf{4.390}$  & $\mathbf{3.172}$  & $\mathbf{31.975}$ \\
& 2 bins + SELayer  & $\mathbf{0.821}$  & $0.534$            & $2.378$           & $25.995$             & $\mathbf{1.191}$  & $5.590$           & $4.333$           & $45.396$              & $\mathbf{1.077}$  & $4.805$           & $3.549$           & $36.822$          \\ \cline{2-14}
& 4 bins            & $0.969$           & $1.781$            & $3.023$           & $36.085$             & $1.505$           & $11.63$           & $4.870$           & $49.077$              & $1.507$           & $12.97$           & $4.652$           & $43.267$          \\     
& 4 bins + SELayer  & $\mathbf{0.844}$  & $\mathbf{0.634}$   & $\mathbf{2.330}$  & $\mathbf{24.777}$    & $\mathbf{1.213}$  & $\mathbf{6.057}$  & $\mathbf{4.322}$  & $\mathbf{44.769}$     & $\mathbf{1.070}$  & $\mathbf{4.625}$  & $\mathbf{3.588}$  & $\mathbf{36.442}$ \\ \cline{2-14}
& 8 bins            & $\mathbf{0.881}$  & $\mathbf{0.672}$   & $\mathbf{2.290}$  & $\mathbf{24.203}$    & $1.292$           & $\mathbf{6.594}$  & $\mathbf{3.978}$  & $\mathbf{42.230}$     & $1.181$           & $5.389$           & $\mathbf{3.346}$  & $\mathbf{33.951}$ \\
& 8 bins + SELayer  & $0.905$           & $0.885$            & $2.308$           & $24.597$             & $\mathbf{1.286}$  & $6.761$           & $4.046$           & $44.366$              & $\mathbf{1.177}$  & $\mathbf{5.318}$  & $3.391$           & $35.452$          \\
\end{tabular}%
}

\end{table*}

\subsection{Squeeze-and-Excitation Layer}
\label{sec:selayer}

Optical flow prediction is a complex task that requires neural networks to extract accurate features precisely describing motion inside the scene. An event aggregation mechanism is therefore required to extract rich temporal features from the events. In Section $4.2$ of the paper we show that time resolution is a key factor for extracting effective feature with Matrix-LSTM. In particular, increasing the number of bins has great impact on the predicted flow and allows the network to retain temporal information over long sequences. Here we focus, instead, on the effect of correlating temporal features by adding a SELayer to the Matrix-LSTM output. Table \ref{tab:opticalflow_SELayer} reports the performance obtained using this additional layer on the MVSEC \cite{zhu2018multivehicle} task. The results we obtained show that adding an SELayer only improves performance on the $4$ bins configuration for the \textit{dt=4} benchmark, while it consistently helps reducing the \textit{AEE} metric on the \textit{dt=1} setting.

By comparing features obtained from subsequent intervals, the SELayer adaptively recalibrates features and helps modelling interdependencies between time instants, which is crucial for predicting optical flow. We believe that a similar approach can also be applied to other event aggregation mechanisms based on voxel-grids of temporal bins to improve their performance, especially those employing data driven optimization mechanisms \cite{gehrig2019end}.

\section{Qualitative Results}
\label{subsec:feature_viz}
The event aggregation process performed by the Matrix-LSTM layer is incremental. Events in each pixel location are processed sequentially; state and output of the LSTM are updated each time. We propose to visualize the Matrix-LSTM surface as an RGB image by using the ResNet18-Ev2Vid configuration and interpreting the $3$ output channels as RGB color. A video of such visualization showing the incremental frame reconstruction on N-Caltech101 samples is provided at this url: \url{https://marcocannici.github.io/matrixlstm}. 

We use a similar visualization technique to show optical flow predictions for \textit{indoor\_flying} sequences. Since we use our best performing model that uses $2$ temporal bins, we decide to only show the first $3$ channels of each temporal interval. Moreover, instead of visualizing how the event representation builds as new events arrive, we only show the frame obtained after having processed each window of events.

%% file: data/plot_chan_vs_batch_time.tex

  \begin{tikzpicture}
  \begin{axis}[
      enlargelimits=false,
  axis y line=right,
  axis x line=none,
  yticklabel pos=right,
      legend pos=north east,
      ylabel={K events/s},
      xlabel={batch size},
       xmin=1, xmax=256,
       ymin=0, ymax=6000,
       symbolic x coords={1,2,8,16,32,64,128,256},
      xtick={1,2,8,16,32,64,128,256},
      ytick={1000,2000,3000,4000,5000,6000},
      ymajorgrids=true,
      grid style=dashed
  ]
  \addplot[
      dashed,
  mark=+,
      color=MaterialGreen700]
  table[x index=0,y index=1,col sep=comma]
  {data/chan_vs_batch_1chan_kev.txt};
  \addplot[
      dashed,
  mark=x,
      color=MaterialBlue700]
  table[x index=0,y index=1,col sep=comma]
  {data/chan_vs_batch_2chans_kev.txt};
  \addplot[
      dashed,
  mark=asterisk,
      color=MaterialPurple700]
  table[x index=0,y index=1,col sep=comma]
  {data/chan_vs_batch_3chans_kev.txt};
  \addplot[
      dashed,
  mark=triangle,
      color=MaterialOrange700]
  table[x index=0,y index=1,col sep=comma]
  {data/chan_vs_batch_4chans_kev.txt};
  \addplot[
      dashed,
  mark=diamond,
      color=MaterialRed700]
  table[x index=0,y index=1,col sep=comma]
  {data/chan_vs_batch_8chans_kev.txt};
  \addplot[
      dashed,
  mark=square,
      color=MaterialPink700]
  table[x index=0,y index=1,col sep=comma]
  {data/chan_vs_batch_16chans_kev.txt};
  
  \end{axis}
  \begin{axis}[
      enlargelimits=false,
      ylabel={Mean Time (ms)},
      legend pos=north west,
      xlabel={batch size},
       xmin=1, xmax=256,
       ymin=0, ymax=12,
       symbolic x coords={1,2,8,16,32,64,128,256},
      xtick={1,2,8,16,32,64,128,256},
      ytick={0,2,4,6,8,10,12},
      ymajorgrids=true,
      grid style=dashed,
      legend style={font=\tiny},
      legend style={at={(0.18,0.795)},anchor=west},
      legend cell align={left},
      legend entries={1 channel,
                      2 channels,
                      3 channels,
                      4 channels,
                      8 channels,
                      16 channels}
  ]
  \addplot[
  mark=+,
      color=MaterialGreen700]
  table[x index=0,y index=1,col sep=comma]
  {data/chan_vs_batch_1chan_time.txt};
  \addplot[
      mark=x,
      color=MaterialBlue700]
  table[x index=0,y index=1,col sep=comma]
  {data/chan_vs_batch_2chans_time.txt};
  \addplot[
      mark=asterisk,
      color=MaterialPurple700]
  table[x index=0,y index=1,col sep=comma]
  {data/chan_vs_batch_3chans_time.txt};
  \addplot[
  mark=triangle,
      color=MaterialOrange700]
  table[x index=0,y index=1,col sep=comma]
  {data/chan_vs_batch_4chans_time.txt};
  \addplot[
  mark=diamond,
      color=MaterialRed700]
  table[x index=0,y index=1,col sep=comma]
  {data/chan_vs_batch_8chans_time.txt};
  \addplot[
  mark=square,
      color=MaterialPink700]
  table[x index=0,y index=1,col sep=comma]
  {data/chan_vs_batch_16chans_time.txt};

  \end{axis}
  \end{tikzpicture}

%% file: data/plot_bin_vs_batch_time.tex

  \begin{tikzpicture}
  \begin{axis}[
      enlargelimits=false,
  axis y line=right,
  axis x line=none,
  yticklabel pos=right,
      legend pos=north east,
      ylabel={K events/s},
      xlabel={batch size},
       xmin=1, xmax=256,
       ymin=0, ymax=6000,
       symbolic x coords={1,2,8,16,32,64,128,256},
      xtick={1,2,8,16,32,64,128,256},
      ytick={1000,2000,3000,4000,5000,6000},
      ymajorgrids=true,
      grid style=dashed,
  ]
  \addplot[
      dashed,
      color=MaterialGreen700,
      mark=+]
      table[x index=0,y index=1,col sep=comma]
      {data/bin_vs_batch_1bin_kev.txt};
  \addplot[
      dashed,
      color=MaterialBlue700,
      mark=x]
      table[x index=0,y index=1,col sep=comma]
      {data/bin_vs_batch_2bins_kev.txt};
  \addplot[
      dashed,
      color=MaterialPurple700,
      mark=asterisk]
      table[x index=0,y index=1,col sep=comma]
      {data/bin_vs_batch_4bins_kev.txt};
  \addplot[
      dashed,
      color=MaterialOrange700,
      mark=triangle]
      table[x index=0,y index=1,col sep=comma]
      {data/bin_vs_batch_9bins_kev.txt};
  \addplot[
      dashed,
      color=MaterialRed700,
      mark=diamond]
      table[x index=0,y index=1,col sep=comma]
      {data/bin_vs_batch_16bins_kev.txt};
  
  \end{axis}
  \begin{axis}[
      enlargelimits=false,
      ylabel={Mean Time (ms)},
      legend pos=north west,
      xlabel={batch size},
       xmin=1, xmax=256,
       ymin=0, ymax=12,
       symbolic x coords={1,2,8,16,32,64,128,256},
      xtick={1,2,8,16,32,64,128,256},
      ytick={0,2,4,6,8,10,12},
      ymajorgrids=true,
      grid style=dashed,
      legend style={font=\tiny},
      legend style={at={(0.18,0.82)},anchor=west},
      legend cell align={left},
      legend entries={1 bin,
                      2 bins,
                      4 bins,
                      9 bins,
                      16 bins}
  ]
  \addplot[
      color=MaterialGreen700,
      mark=+]
      table[x index=0,y index=1,col sep=comma]
      {data/bin_vs_batch_1bin_time.txt};
  \addplot[
      color=MaterialBlue700,
      mark=x]
      table[x index=0,y index=1,col sep=comma]
      {data/bin_vs_batch_2bins_time.txt};
  \addplot[
      color=MaterialPurple700,
      mark=asterisk]
      table[x index=0,y index=1,col sep=comma]
      {data/bin_vs_batch_4bins_time.txt};
  \addplot[
      color=MaterialOrange700,
      mark=triangle]
      table[x index=0,y index=1,col sep=comma]
      {data/bin_vs_batch_9bins_time.txt};
  \addplot[
      color=MaterialRed700,
      mark=diamond]
      table[x index=0,y index=1,col sep=comma]
      {data/bin_vs_batch_16bins_time.txt};

  \end{axis}
  \end{tikzpicture}